\documentclass{article}

\usepackage{microtype}
\usepackage{graphicx}
\usepackage{subfigure}
\usepackage{booktabs}
\usepackage{xurl}
\usepackage{hyperref}

\usepackage[accepted]{icml2025}

\makeatletter
\renewcommand{\Notice@String}{Preprint.}
\makeatother

\hypersetup{
    pdftitle={WEIGHTBENCH: Evaluating Language Models as Weight-Space Programmers},
    pdfauthor={Aayush Marasini, Barsat Khadka, Nick Rahimi},
    pdfsubject={Preprint},
    pdfkeywords={Weight-Space Programming, Large Language Models,
                 Neural Network Editing, Benchmarking}
}

\usepackage{amsmath}
\usepackage{amssymb}
\usepackage{mathtools}
\usepackage{amsthm}
\usepackage[capitalize,noabbrev]{cleveref}

\newcommand{\WB}{\textsc{WEIGHTBENCH}}

\theoremstyle{plain}

\theoremstyle{definition}

\theoremstyle{remark}

\usepackage[textsize=tiny]{todonotes}

\icmltitlerunning{WEIGHTBENCH}

\begin{document}

\twocolumn[

\icmltitle{WeightBench: Evaluating Language Models as Weight-Space Programmers}


\icmlsetsymbol{equal}{*}

\begin{icmlauthorlist}

\icmlauthor{Aayush Marasini}{equal,usm}

\icmlauthor{Barsat Khadka}{equal,usm}

\icmlauthor{Nick Rahimi}{usm}

\end{icmlauthorlist}


\icmlaffiliation{usm}{
The University of Southern Mississippi,
Hattiesburg, MS 39401, USA
}


\icmlcorrespondingauthor{Barsat Khadka}{Barsat.Khadka@usm.edu}


\icmlkeywords{
Weight-Space Programming,
Large Language Models,
Neural Network Editing,
Benchmarking
}

\vskip 0.3in

]


\printAffiliationsAndNotice{\icmlEqualContribution}


\begin{abstract}

Recent work shows that large language models (LLMs) can operate
directly on the parameters of other neural networks.
They can \emph{read} what a network was trained to do,
\emph{fix} a network that fails, and \emph{write} working
networks from scratch, all without gradient descent.
This capability has been termed
\emph{Weight-Space Programming} (WSP). Exploring this new paradigm systematically requires a
robust benchmark. We introduce \WB{}, a benchmark that
evaluates LLMs of increasing capability on 42 WSP tasks
across five tracks: reading, fixing and writing weights,
recognizing edits that are impossible, and data-free
compression. The networks whose weights the LLMs read and edit range
from $2K$-parameter control policies to $7B$-parameter
language models. Experimental benchmark results show that frontier agents can already read small networks and write working control policies without ever running them, but cannot yet patch language models, compress them without data, or tell an impossible edit from a hard one. On all 42 tasks, Claude Opus 5.5 reaches a macro score of 0.535 (Sonnet 5.5 0.466, Haiku 5.5 0.402, DeepSeek v4-flash 0.215); on the 38 tasks it was run on, GPT-6.1-sol reaches 0.568 against Opus's 0.546. Code, the task registry, our OS-level sandbox harness, and the live leaderboard are publicly available at \url{https://github.com/Aayush-Marasini/WeightBench}.

\end{abstract}


\section{Introduction}

Traditionally, neural networks are developed by specifying
an architecture and learning its parameters through
optimization. Once trained, these weights are generally
treated as the output of the learning process rather than
something that can be read or programmed directly.
However, previous work on model editing has shown that
specific behaviors can be changed through targeted
modifications to network weights
\citep{locatingandediting}, while compilers such as
Tracr have demonstrated that Transformer weights can
be constructed directly from symbolic programs
\citep{Tracr}. More recently, \citet{takehi2026weightspace}
demonstrated that large language models (LLMs) can perform
a broader range of operations on neural network weights
themselves. They showed that LLMs can identify what a
network was trained to do, repair failures by modifying
its parameters, and construct working networks from
scratch without gradient descent. This capability has
been termed \emph{Weight-Space Programming} (WSP),
drawing an analogy to traditional programming, but
operating directly on neural network weights rather
than source code.

Although these demonstrations show that WSP is possible,
they do not yet establish how reliably LLMs can perform
these operations across different networks and tasks.
For instance, identifying the behavior of a small control
policy is different from locating learned behaviors in
a large language model, and successfully repairing one
failure does not necessarily mean that other capabilities
are preserved. It is also unclear how these abilities
change with the capability of the LLM performing the
weight-space operations. Understanding these differences
requires a systematic evaluation across network
architectures, model sizes, and task difficulties.

Evaluating WSP also presents challenges that are not
typically encountered in conventional language model
benchmarks. An agent may inspect weight tensors, analyze
them using numerical tools, or execute the network to
observe its behavior. These forms of access are not
equivalent. With unrestricted forward passes, an agent
could discover successful edits through repeated trial
and error rather than reasoning about the weights.
Furthermore, an edit that improves a target behavior
may degrade unrelated capabilities, while some requested
modifications may be infeasible under the given
constraints. A meaningful evaluation must therefore
control how agents access and execute networks, while
measuring both the intended improvements and their costs.

To systematically evaluate these capabilities, we
introduce \WB{}, a benchmark for evaluating LLMs as
weight-space programmers. \WB{} defines 42 task
specifications across five tracks: reading weights,
repairing networks, constructing networks from scratch,
recognizing infeasible modifications, and data-free
compression. The tasks span networks ranging from
2K-parameter control policies to 7B-parameter
language models. We evaluate agents under explicit
access and compute constraints, with task-specific
scoring that accounts for performance improvements,
unintended behavioral changes, and the costs of
modifications. Non-agentic baselines and reference
solutions are included to establish task difficulty
and measure the available room for improvement.

\section{Background}

\subsection{Weight-Space Programming}

Weight-Space Programming (WSP), introduced by
\citet{takehi2026weightspace}, investigates whether
general-purpose LLMs can directly analyze and manipulate
the parameters of neural networks. Rather than relying on
a fixed editing algorithm, an agent examines the weights,
performs numerical analysis when needed, and determines
how to modify or construct the network. The initial
demonstrations include identifying the tasks of trained
networks, repairing failing policies, and constructing
game-playing networks without gradient descent.
Importantly, these experiments involve different levels
of access, ranging from raw weights alone to code-assisted
analysis and limited behavioral feedback.

\subsection{Weight-Space Learning}

Weight-Space Learning (WSL) is a broader research direction
that treats neural network parameters as structured data
that can be analyzed, represented, and generated.
\citet{han2026surveyweightspacelearning} categorize WSL
into three areas: weight-space understanding, representation,
and generation. Weight-space understanding studies
properties such as parameter symmetries, where different
weight configurations can represent the same function.
Weight-space representation learns useful features from
model parameters, with methods such as Deep Weight-Space
Networks \citep{pmlr-v202-navon23a} and Neural Functional
Networks \citep{NEURIPS2023_4e9d8aee} accounting for
neuron permutation symmetries. Weight-space generation
extends this idea to constructing model parameters through
hypernetworks \citep{Ha2017} and generative models such
as HyperDiffusion \citep{Erkoc_2023_ICCV}. While WSL
develops specialized methods for understanding and
generating weights, WSP investigates whether
general-purpose LLMs can perform such operations through
reasoning and tool use.

\subsection{LLM Agent Benchmarks}

Recent benchmarks evaluate LLM agents on tasks requiring
tool use, code execution, and multi-step problem solving.
SWE-bench \citep{swebench} evaluates software issue
resolution, MLE-bench \citep{mlebench} examines machine
learning engineering, and PaperBench \citep{paperbench}
evaluates the replication of published research. These
benchmarks measure agents through executable artifacts
and task-specific evaluation criteria. Evaluating WSP
introduces additional constraints, as unrestricted model
execution or optimization can make it difficult to
distinguish weight-level analysis from behavioral search.
WeightBench builds on artifact-based agent evaluation
while focusing specifically on operations performed on
neural network parameters.

\section{Evaluation Framework}

\subsection{Execution Environment and Auditing}
\label{sec:3.1}
The main constraint of WSP is that modifications must  be done from geometric or topological analysis of the neural network weights, not iterative or output-driven optimization. A naive evaluation framework is vulnerable to agents silently running gradient descent or unmetered froward passes to find solutions using train-and-error. To mitigate this, WeightBench boundes and observes an agent's access, compute and edits, and scores every task under a  strict budget. It executes all tasks within a strict, unprivilated OS-level sandbox. The agent is completely isolated from the host filesystem except for the task's public artifacts and its workdir, and is denied access to model execution libraries (such as \texttt{transformers} or \texttt{vllm}), and held in a strict network lockdown where the only exits are its  own LLM API and the local, token-authenticated Weight API. 

We cannot enforce a zero forward pass constraint ($N_forward = 0$) with a naive FLOP cap. Allowed analysis, such as computing the Singular Value Decomposition (SVD) of weight matrices, routinely costs more FLOPs than a forward pass. Therefore, WeightBench injects a metered NumPy subclass into every agent process and traces arithmetic on weights at the operator level. From these traces, our runtime guard detects forward-pas chains and records the violation. 

To guarantee the integrity of the evaluation, every metered forward pass, probe and runtime-guard event is recorded in a hash-chained ledger. If an agent tries to bypass the runtime guards, the system flags it and quarantines the run. The quarantined run scores $0$ until a human review clears it. The final scores are computed directly from this ledger rather than the agent's outputs. Thus we never have to trust the agent's self reported work and making every submitted result verifiable.

\begin{figure}[ht]
    \centering
    \includegraphics[width=\linewidth]{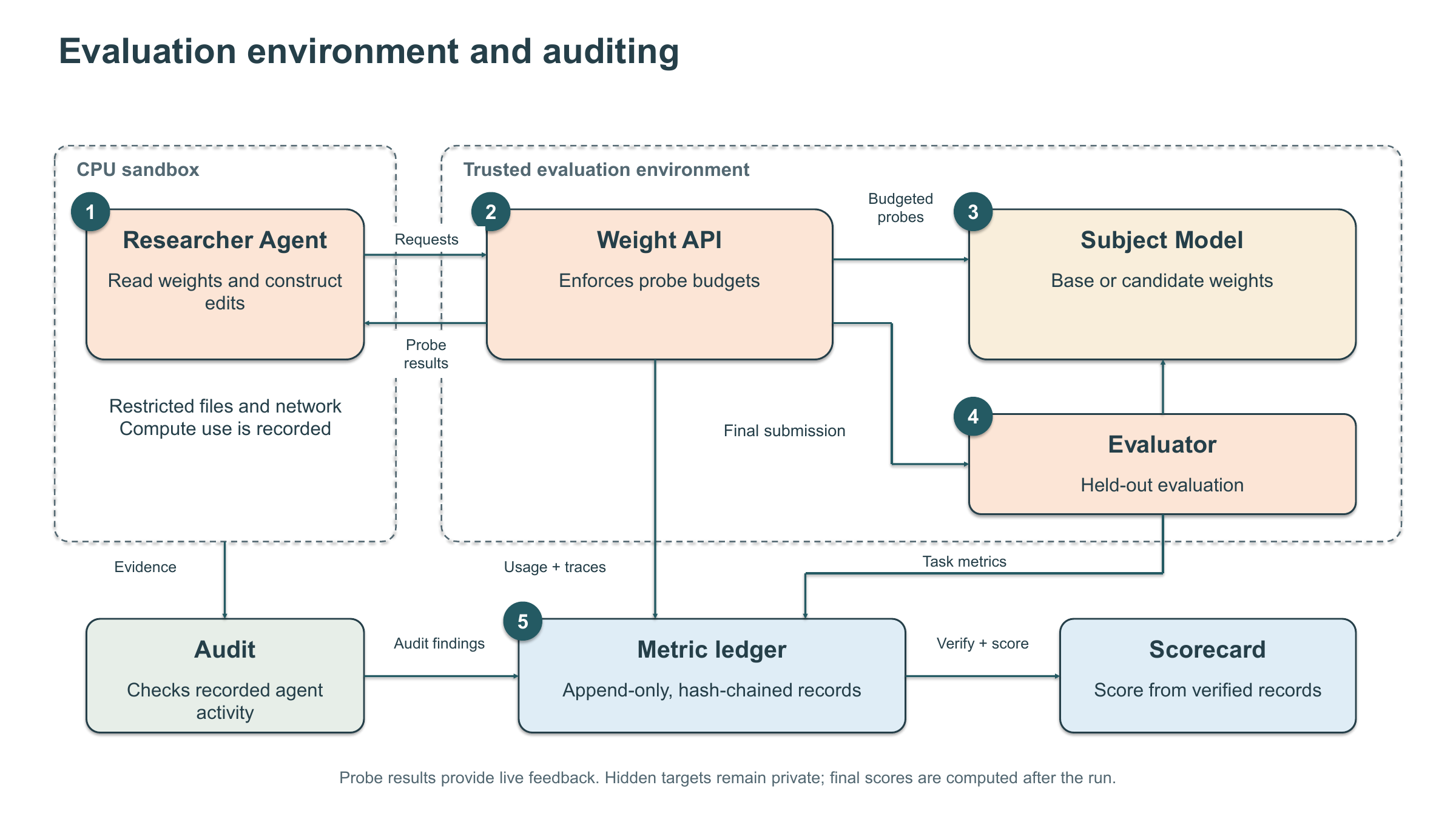}
    \caption{The WeightBench evaluation pipeline.}
    \label{fig:eval_pipeline}
\end{figure}

\subsection{Evaluation Metrics}
To properly evaluate an agent's ability to edit weights, we cannot just look at whether it solved the problem. A successful edit must improve the target capability while minimizing the edit's footprint and computational overhead. It must also do so without causing catastrophic degradation of other capabilities of the neural network. Therefore, WeightBench evaluates the agent using a composite score, $S_{WB}\in [0,1]$, defined by four components:

\begin{itemize}
    \item Task Gain $(S^{+} \in [0,1])$ measures if the agent did the job. It tracks accuracy for tasks involving reading the weights, and for tasks involving diagnosis, the metric calculates how much the agent improved over the base model, divided by the total "room for improvement" available between the base model and the absolute ceiling.
    \item Collateral Damage $(D_{col} \in [0,1])$: quantifies the degradation of other capabilities caused by the weight edits. It is defined as the performance drop of the edited mode on held-out benchmarks (e.g., MMLU). It is lower bounded at $0$ to ensure that accidental improvements do not offset performance losses elsewhere.
    \item {Structural Edit Cost ($u_{\text{edit}} \in [0,1]$):} evaluates how precise and surgical the agent's weight changes are. For tasks that don't use an edit budget (like reading or building from scratch), we set $u_{\text{edit}} = 0$. For tasks that do, it tracks the size of the update using the smaller of the $L_0$ parameter count fraction and the matrix rank fraction. If a submission blows past either budget limit, it gets disqualified ($0$). Ultimately, this metric penalizes broad, messy edits and rewards clean, targeted fixes.
    \item Compute Use $(E/E_{budget} \in [0,1])$ accounts for the computational resources expended. It is measured as the ratio of metered diagnostic probes and forward passes used relative to the task's budget. 
\end{itemize}
We combine these metrics to get a final WeightBench score via the following equation:
\begin{equation}
    S_{\mathrm{WB}} = S^{+} \cdot e^{-\alpha D_{\mathrm{col}}} \cdot (1 - \kappa u_{\mathrm{edit}}) \cdot \left( 1 - \beta \frac{E}{E_{\mathrm{budget}}} \right)
    \label{eq:swb}
\end{equation}
To make hacking a specific metric unprofitable, we utilize fixed constants. We set alpha to $\frac{-\ln 0.8}{0.1} \approx 2.23$ so that a $10\%$ drop on the collateral benchmark reduces the overall score by $20\%$, strictly punishing collateral damage. The constants $\kappa=0.1$ and $\beta=0.1$ ensure that maxing out the allowed edit or compute budgets results in a $10\%$ penalty. And, because the penalties multiply, a small edit cost cannot compensate for a small gain. A useless edit scores near zero however cheap it is, and no factor can be traded against another. Consequently, $S_{\text{WB}} \ge 0.81 S^+ e^{-\alpha D_{\text{col}}}$ for any valid submission, so the edit and compute factors only break ties between working solutions.

\subsection{Contamination Mitigation and Procedural Scaling}
Data contamination is a really big vulnerability for benchmarks. AI agents can memorize the coordinates of known vulnerabilities in open-weight models, across different releases. To limit this, we enforce a temporal refresh. Subject models are regenerated using fresh random seeds, applying rotations and hidden dimension permutations to the architecture. This helps preserve the functional behavior of the network while scrambling its internal geo metry, so that the agents must analyze the weights rather than memorize indices. 

Moreover, to ensure the benchmark scales alongside frontier models getting better, we use exponential difficulty knobs to set the difficulties of the tasks. Each track maps the difficulty level, $L$, to a quantity using the formula $v(L) = v_{0}\times g^{L}$. For example, in a task where the agent must find and remove "dead" nurons, at $L=1$, the agent must trace this logic through $3$ hidden layers, at $L=2$, through $9$ layers, and so on, currenlty capped at $24$ layers. This exponential expansion keeps the benchmark from saturating as models improve.

\section{Benchmark Tracks}
WEIGHTBENCH consists of 42 registered tasks (40 specification tasks plus two synthetic knob families) structured on a $50/48/2$ tier distribution, scaling from foundational $2K$ parameter MLPs to $7B$ parameter Transformers. To ensure every task has a static reference solution, we use public methods for our reference solutions. Every track has a non-agentic baseline and a public static reference solution. The reference beats the baseline by a measured margin on the tasks that are wired end to end. We distribute the tasks into five tracks to isolate different weight-space capabilities ($10$ READ, $13$ FIX, $5$ META, $6$ WRITE and $8$ SYSTEMS tasks).
\subsection{Static Diagnosis (READ)}
READ tasks evaluate whether an agent can understand the properties of a model from its weights only, with $N_{forward}$ set to $0$. We instruct the agents to identify training environments from raw MLP weights, locate induction heads from Transformers that have had their input/output embeddings removed, and to recover the concept a fine-tuning batch taught from a checkpoint by looking at the difference in its weights after fine-tuning. 

\subsection{Surgical Patching (FIX)}
FIX tests if an agent can do surgical weight interventions to fix specific bugs with minimal collateral damage other general capabilities of the subject model. Agents must mathematically project out the directions that cause repetitive degenerate generation in small language models, fix sycophancy  using restricted probing tools, and write override circuits to fix failing physics policies.

\subsection{Infeasibility Verification (META)}
Under tight parameter budgets and collateral preservation constraints, some requested interventions are practically infeasible. Editing weights can remove an obstacle that hides an existing skill (e.g., fixing a loop to recover latent math), but targeted editing fails to inject complex knowledge that a model never learned. We treat edits requiring knowledge the base model never had as infeasible and test meta awareness by giving the agent infeasible tasks mixed with difficult but solvable decoys. A successful agent will diagnose between the blocked skills and the infeasible skills, and refuse to edit weights for the impossible task.

\subsection{Zero Shot Weight Synthesis (WRITE)}
This track tests if an agent can synthesize neural networks from scratch. It starts from a blank slate of all zeros andf without using training loops or gradient descent, the agent must write the exact weight numbers required to play video games like Breakout or Go. We also task agents to move a working logic circuit from a simple network into a larger language model. 

\subsection{Hardware Aware Optimization (SYSTEMS)}
Finally, this track tests if an agent can compress a model to run on small edge devices. The agent must delete unnecessary connections and rebalance the remaining weights using pure math, without running any calibration. It must prove dead neurons can never fire and prune them. Agents must enforce hardware-friendly limits and and perform outlier migration so the model can run efficiently on low-power chips. 

\section{Experiments}
The current research climate poses a fundamental question: Is Weight-Space Programming (WSP) a genuine, emerging reasoning capability, or just an illusion driven by brute-force trial-and-error and training data contamination? To answer this, our evaluation removes all the safety nets of traditional optimization. We investigate whether current Large Language Models can truly operate as weight-space programmers when forced to rely strictly on geometric and topological reasoning under severe access and compute constraints.

We evaluate the agents across five tracks, measure how their capabilities degrade as task complexity scales, and analyze the failure modes. What we found was that agents are already good at reading small networks, and the strongest models can write a working game-playing network from zero without ever running it. But, no agent came close to a gradient-fitted fix when asked to repair a language model. The models were also particularly bad at compressing a model well without data and at differentiating an infeasible edit from just difficult ones.

\begin{table*}[t]
\centering
\small
\setlength{\tabcolsep}{6pt}
\begin{tabular}{@{}l ccc c c@{}}
\toprule
 & \multicolumn{3}{c}{Claude 5.5} & & \\
\cmidrule(lr){2-4}
Track (tasks) & Haiku & Sonnet & Opus & DeepSeek v4-flash & GPT-6.1-sol$^{\dagger}$ \\
\midrule
TR-01 \textsc{Read}: Static Diagnostics (10)          & 0.671 & 0.656 & 0.744 & 0.470 & 0.928 \\
TR-02 \textsc{Fix}: Surgical Patching (13)            & 0.300 & 0.381 & 0.404 & 0.187 & 0.571$^{\dagger}$ \\
TR-03 \textsc{Meta}: Infeasibility Verification (5)   & 0.491 & 0.500 & 0.492 & 0.181 & 0.260 \\
TR-04 \textsc{Write}: Zero-Shot Weight Synthesis (6)  & 0.369 & 0.569 & 0.829 & 0.000 & 0.863 \\
TR-05 \textsc{Systems}: Hardware-Aware Optimization (8) & 0.177 & 0.225 & 0.207 & 0.238 & 0.220 \\
\midrule
\textbf{Macro $S_{\mathrm{WB}}$ (42 tasks)}           & 0.402 & 0.466 & \textbf{0.535} & 0.215 & -- \\
Macro $S_{\mathrm{WB}}$ (38 tasks, no TR-02 loops)   & 0.428 & 0.491 & 0.546 & 0.232 & \textbf{0.568} \\
\midrule
Tasks evaluated                                      & 42 & 42 & 42 & 42 & 38 \\
Runs (quarantined)                                   & 78 (7) & 77 (1) & 77 (11) & 42 (13) & 38 (7) \\
\bottomrule
\end{tabular}
\caption{Agent performance on the five \WB{} tracks at the default difficulty level, seed~0: mean $S_{\mathrm{WB}} \in [0,1]$ per track (task scores averaged over runs, then over the track's tasks); the macro score is the mean of the five track means. Claude models ran two samples per task, DeepSeek and GPT-6.1-sol one. Quarantined runs and runs that ended without a valid submission score 0; no quarantined run has been reviewed yet, so all scores are lower bounds. $^{\dagger}$GPT-6.1-sol was not run on the four TR-02 reasoning-loop tasks, so its TR-02 mean covers 9 of 13 tasks and its macro is reported only on the 38 shared tasks; 10 of its 38 runs used the Codex CLI scaffold instead of Claude Code. The evaluation of Claude Fable 5.1 is ongoing and not reported.}
\label{tab:agent-results}
\end{table*}

\subsection{Experimental Setup}
We evaluated five models on the 42 tasks of the \WB{} registry. Claude Opus 5.5, Claude Sonnet 5.5, Claude Haiku 5.5 and DeepSeek v4-flash were run on all 42 tasks. GPT-6.1-sol was run on 38: all except the four TR-02 reasoning-loop tasks, each of which needs one to one and a half GPU-hours per run. The evaluation of Claude Fable 5.1 is ongoing; its partial results are not reported here and will be included in a later version.

All agents interact with our subject models strictly through the isolated CPU sandbox described in Section~\ref{sec:3.1}. We route the agents through a headless adapter that enforced our strict budget parameters. In every run, agents are asked to solve one task, and they only receive the weights, the task specification and the metered probes the task allows. All runs evaluate the default difficulty level on a fixed seed (seed~0), with two samples per task for the Claude models and one for DeepSeek and GPT-6.1-sol, all from one frozen version of the harness. DeepSeek and GPT-6.1-sol were served through OpenRouter and ran with an 80-turn cap. A run that ends without a valid submission, including one that reaches the turn cap, scores 0; only infrastructure failures (a crashed agent CLI, an exhausted API credit, a full disk, an evaluator crash before a result was written) were re-run under a fresh sample identifier.

Crucially, all the scores are computed from the run's cryptographically hashed ledger and not taken from the agent's own report. Every run is subjected to an automated audit scanning for $28$ different quarantine flags (e.g., gradient descent attempts, sandbox escapes, etc). If our audit flags a run, it is quarantined and scores $0$ until a human reviews it. None of the quarantined runs has been reviewed yet, so every score we report is a lower bound. A task's score is computed as a mean of a model's runs, and a track's score is the mean of all of its tasks, and finally, the macro score is the average over all five tracks.

\subsection{Task Validity and Baselines}
Before evaluating the LLM agents, we need to establish that a task is actually solvable without relying on trial and error optimization. We first validated all $42$ tasks against non-agentic baselines. 

For the tasks to be valid, a reference solution had to achieve a WeightBench score $S_{\mathrm{WB}}$ of at least $0.5$ and beat the naive baseline by a margin of at least $0.2$. Currently $39$ out of the $42$ tasks clear this validity gate at their default difficulty levels. However, all of the reference solutions are not equal. For 21 of the 42 tasks, the reference solution only uses what the agent has access to, and for the rest, the reference solution requires privileged access, such as gradient fitting on calibration prompts. A privileged reference proves that a solution exists within the weights, but does not guarantee it can be found using only data-free analysis.

\subsection{Agent Performance}
Table~\ref{tab:agent-results} has the macro WeightBench scores across the five tracks. These results show a clear scaling behaviour among the fully evaluated models. Ordered by size, Claude Opus 5.5 ($0.535$) outperforms Sonnet ($0.466$), which outperforms Haiku ($0.402$). DeepSeek v4-flash struggled under the strict scaffolding and auditing constraints: it has the lowest overall score ($0.215$) and the highest quarantine rate (13 of 42 runs). GPT-6.1-sol, run on 38 tasks, has the highest score on the tasks it shares with the other models ($0.568$ against Opus's $0.546$), with the best TR-01 ($0.928$) and TR-04 ($0.863$) means, but it scores well below the Claude models on TR-03 ($0.260$).

Our different tracks differentiate agent capabilities very unevenly. TR-04 (Weight Synthesis) separated the models the most. Opus reached $0.829$ when synthesizing weights zero-shot from scratch, against Sonnet ($0.569$), Haiku ($0.369$) and DeepSeek ($0.000$); GPT-6.1-sol reached $0.863$. But in TR-03 (Infeasibility Verification) the three Claude models performed alike, between $0.485$ and $0.500$ on every task, which is essentially the baseline of random guessing. This indicates that current models find it difficult to distinguish between a blocked capability and a missing capability by analyzing weights only. 

All agents performed poorly on TR-05 (Hardware-Aware Optimization), between $0.177$ and $0.238$. This also confirms our reference results that non-privileged, data-free compression remains a difficult challenge. Finally, Surgical Patching (TR-02) remains largely open. Opus is a small step ahead of the other fully evaluated models, but the agents largely failed to execute targeted behavioral fixes without collateral damage to the subject model's general capabilities. On the four reasoning-loop tasks, only Opus (on three) and Sonnet (on one) scored above zero.

\section{Conclusions}
Weight-Space Programming represents a shift in how we interact with trained neural networks, treating them not just as static artifacts, but as targets that can be programmed. We introduced WeightBench to systematically measure whether LLMs can genuinely read, fix, and write neural network weights using only geometric and topological reasoning.

Our results provide valuable insights into the current debate if WSP is just a trick or byproduct of data contamination or if LLMs can actually be weight space programmers. The ability of models to synthesize functional control policies from a blank slate, entirely zero-shot, confirm that LLMs can map natural language semantics directly to geometry of neural network parameters. But, our results also reveal that this capability is in its infant phase. Models still lack the precision required to surgically patch issues or safely compress massive language models without triggering catastrophic collateral damage. 

By strictly auditing compute, isolating the execution environment, and enforcing hard edit budgets, WeightBench prevents agents from bypassing WSP constraints through brute-force trial and error. As we continue to expand the benchmark to include complete evaluations of models like Claude Fable and explore dynamic difficulty scaling, we intend WeightBench to serve as the rigorous proving ground necessary to mature Weight-Space Programming from an emergent curiosity into a reliable engineering discipline.

\section*{Acknowledgments}
This work used the high-performance computing resources of the Mississippi Center for Supercomputing Research (MCSR). The authors thank MCSR for providing this supercomputing access and its staff for their support. 

\bibliography{example_paper}
\bibliographystyle{icml2025}



\newpage
\appendix
\onecolumn
\newpage
\appendix
\onecolumn

\section{Execution Sandbox and Guard Injection}
\label{app:sandbox}
The OS-level isolation and runtime guard operate independently. The OS sandbox bounds what the process can reach, while the runtime guard meters what it computes. 

On Linux, the backend utilizes \texttt{bubblewrap} with isolated user, PID, IPC, UTS, cgroup, and network namespaces (\texttt{--unshare-all}), alongside a private \texttt{/proc} and \texttt{/dev}. Agent processes inherit a hard CPU-time \texttt{rlimit} equal to the task's wall-clock budget and a maximum file size constraint. The agent is denied write access outside its designated working directory, and the execution of compilers or foreign interpreters is blocked. 

The runtime guard is injected through Python's site machinery. Because agent CLIs frequently execute analysis scripts using the \texttt{-I} flag (which ignores \texttt{PYTHONPATH}), injecting the guard solely via environment variables leaves the system vulnerable to unmetered execution. To prevent this, the harness installs the guard through a \texttt{.pth} hook in the agent environment's read-only \texttt{site-packages}, which Python processes even under \texttt{-I} or \texttt{-E}. The only flag that bypasses the site machinery entirely is \texttt{-S}. The audit detects \texttt{-S} in all its forms (shell word, argv-list element, or combined \texttt{-IS}) and quarantines the run as a \texttt{guard\_bypass\_attempt}.

\section{Operator Metering and Data-Flow Provenance}
\label{app:metering}
To enforce FLOP caps without preventing static analysis, every weight array loaded from disk is wrapped in a metered NumPy subclass. Each matrix multiplication charges $2mkn$ FLOPs to the process. 

Because forward-pass chains (a weight matrix applied to activations, followed by a nonlinearity and another weight matrix) must be distinguished from legitimate mathematical analysis, chain detection follows data-flow provenance rather than static shape heuristics. Each metered array carries a tag. A product involving a weight whose values were modified by a nonlinearity (e.g., a sign split, absolute value, or masked weight) is classified as bound propagation rather than a network layer. Consequently, interval-bound propagation of the form $\max(W,0)u + \min(W,0)l$ is not incorrectly penalized as a forward pass. Elementwise arithmetic is not FLOP-charged.

\section{Audit Flags and Quarantine Logic}
\label{app:audit}
The post-run audit evaluates three evidence sources: the agent transcript, the runtime-guard traces, and the egress-proxy log. If a run raises any unpermitted flag, it is quarantined and scores zero until cleared by manual review. The 28 quarantine flags are divided into four layers:

\begin{itemize}
    \item \textbf{Text Rules:} Scans agent-written code for evasion or unauthorized execution. Flags: \texttt{model\_execution\_library}, \texttt{simulator\_execution}, \texttt{gradient\_descent}, \texttt{guard\_bypass\_attempt}, \texttt{network\_attempt}, \texttt{other\_interpreter}.
    \item \textbf{Runtime Rules:} Triggered actively by the guard during execution. Flags: \texttt{blocked\_import}, \texttt{blocked\_gradient}, \texttt{blocked\_network}, \texttt{egress\_denied}, \texttt{repo\_path}, \texttt{metering\_escape}, \texttt{raw\_weight\_read}, \texttt{unsupported\_contraction}, \texttt{forward\_pass\_suspect}, \texttt{forward\_chain\_runtime}, \texttt{flop\_cap\_exceeded}, \texttt{compute\_budget\_exceeded}, \texttt{trace\_limit\_exceeded}.
    \item \textbf{Integrity Rules:} Detects missing, malformed, or tampered evidence. Flags: \texttt{no\_transcript}, \texttt{no\_runtime\_trace}, \texttt{trace\_tampered}, \texttt{trace\_malformed}, \texttt{transcript\_malformed}, \texttt{egress\_malformed}, \texttt{agent\_root\_spoof}.
    \item \textbf{Accounting Rules:} Validates CPU subtree agreement. Flags: \texttt{unguarded\_cpu}, \texttt{cpu\_tree\_unreported}.
\end{itemize}

\section{Red-Team Audit Validation}
\label{app:redteam}
To validate the efficacy of the harness and audit flags, we executed a scripted red-team array containing 29 cheat scenarios (e.g., attempting gradient descent, raw weight reads, batch-of-one forward chains, and sandbox escapes) and 30 honest controls. 

Evaluated over three independent trials per scenario on the cluster (180 total runs), the audit successfully caught and quarantined 28 of the 29 cheats in all three trials. The single unflagged cheat (\texttt{no\_transcript}) was successfully blocked by the OS sandbox, preventing the exploit despite not raising a specific trace flag. The system recorded 0 false positives across all 90 honest control trials. Because every scenario behaved identically across all trials, the scripted agent's detection rate is deterministic. 

\section{Generators and Anti-Memorization}
\label{app:generators}
To ensure agents perform mechanistic weight-space reasoning rather than memorizing known vulnerabilities from public training data, subject models are generated procedurally from a seed. Before shipment to the agent, the generator applies seeded orthogonal rotations and hidden-dimension permutations to the architecture. Only permutations are utilized where rotations are not equivariant (e.g., LayerNorm with per-coordinate gains). The transformed model computes the identical mathematical function, but the stored coordinates are entirely scrambled, invalidating any memorized indices.
\end{document}